\documentclass[preprint, 12pt]{elsarticle}

\usepackage{amsmath,amssymb,amsfonts}
\usepackage{textcomp}
\usepackage{hyperref}
\usepackage{amsmath}
\usepackage{amssymb}
\usepackage{graphicx}
\usepackage{float}
\usepackage{algorithm}
\usepackage{algpseudocode}
\graphicspath{ {./images/} }
\usepackage{comment}

\begin{document}

\begin{frontmatter}

\title{Leveraging KANs For Enhanced Deep Koopman Operator Discovery}

\author{George Nehma and Madhur Tiwari} 

\affiliation{organization={Department of Aerospace, Physics and Space Sciences, Florida Institute of Technology},
            addressline={150 W University Blvd}, 
            city={Melbourne},
            postcode={32901}, 
            state={FL},
            country={USA}}

\begin{abstract}

Multi-layer perceptrons (MLP's) have been extensively utilized in discovering Deep Koopman operators for linearizing nonlinear dynamics. With the emergence of Kolmogorov-Arnold Networks (KANs) as a more efficient and accurate alternative to the MLP Neural Network, we propose a comparison of the performance of each network type in the context of learning Koopman operators with control. In this work, we propose a KANs-based deep Koopman framework with applications to the pendulum, the combined pendulum-cart system and an orbital Two-Body Problem (2BP) for data-driven discovery of linear system dynamics. KANs were found to be superior in nearly all aspects of training; learning 31 times faster, being 15 times more parameter efficiency, and predicting 1.25 times more accurately as compared to the MLP Deep Neural Networks (DNNs) in the case of the 2BP. Thus, KANs shows potential for being an efficient tool in the development of Deep Koopman Theory.
\end{abstract}
\end{frontmatter}

\section{Introduction} \label{sec:I}
The development of Koopman Theory for the purpose of improving the linearization of nonlinear systems, system identification and the development of control systems has garnered great interest over the last few years. The use of Multi-Layer Perceptron (MLP) Deep Neural Networks (DNNs) to aid in the discovery of the Koopman operator has shown to have great performance and potential \cite{Tiwari2023-ya,Nehma2024-fi}. However, a main drawback to this approach is the often tedious and prolonged training cycles that must be iterated over to learn a stable and accurate approximation to the Koopman operator. Kolmogorov-Arnold Networks (KANs), with their deep network architecture as in \cite{Liu2024-dx}, offer a prospective alternative to MLP DNNs, with the promise of improved accuracy, shorter training times, and less training data required. 

The use of the Kolmogorov-Arnold theorem in the development of neural networks has been studied previously \cite{Sprecher2002-xg, Fakhoury2022-yp, Lin1993-iz} but in all of these cases the network was studied with a fixed depth and width, resulting in a network that cannot be considered deep. However, the recent work in \cite{Liu2024-dx} has shown promising improvements to KANs by addressing some of the major issues with the original framework. In \cite{Liu2024-dx} the authors were able to demonstrate a number of applications using KANs in which the model was both more efficient and accurate than the MLP counterpart. 
This work is an extension of our previous works \cite{Tiwari2023-ya, Nehma2024-fi} where we proposed a deep learning framework using the traditional MLPs for efficient learning of Koopman operators. Here, we modify the framework using deep KANs and show that this new framework significantly improves the existing architecture used in conjunction with Koopman theory to create a more efficient and accurate learning framework for linearizing nonlinear dynamics for the purpose of control, state estimation, and more. We also introduce the application of our KANs architecture on a more nonlinear, complex and fully under actuated system in the pendulum-cart system to test its ability to efficiently and accurately develop a nonlinear controller.

Koopman theory, first proposed in 1931 \cite{Koopman1931-ev}, has recently been viewed as a solution for improved linearization of nonlinear systems. In a nutshell, the theory states that an infinite-dimensional linear Koopman operator can exactly describe the dynamics of a system. Due to the impractical limitation of infinite dimensions in reality, it is approximated using data-driven methods such as Extended Dynamic Mode Decomposition (EDMD) \cite{Korda2018-qs,Lusch2018-oh} or EDMD with control (EDMDc) \cite{Folkestad2020-yk} for the implementation on a controlled system. State-of-the-art deep Koopman theory adopts a feed-forward neural network to make it possible to find an approximate general Koopman operator linearization applicable to a larger region of state space. One of the main advantages of using a data-driven methods is that they do not require system knowledge, and the nonlinear system can be completely unknown.

However, a shortcoming of this method is that large training data requirements, especially in real-world scenarios \cite{Zhan2023-xs,Manzoor2023-md, Zhang2023-hn}, could be lengthy, difficult, or not possible at all. Furthermore, the more complex the dynamics of the system, the larger the required network is to learn and find an adequate approximation, reducing the overall efficiency of the method. But, with improvements in neural networks, it is hoped that models such as KANs can alleviate these fundamental training issues. KANs is a new network architecture that is designed to be more efficient because of the way it approximates any nonlinear function. Whereas traditional MLPs utilize the Universal Approximation Theorem (UAT) \cite{Cybenko1989-sd}, KANs takes advantage of the Kolomogorov-Arnold Theorem \cite{Liu2024-dx}, allowing for a more concise and efficient network. On a per node basis, KANs takes more time to train than its MLP counterpart, however the main idea is that a significantly smaller KANs network can outperform a comparable MLP DNN. This is highly advantageous for practical robotics, aerospace and other autonomous systems where computational power, and size is often a limiting design factor. 

To this end, the contributions we propose are three-fold:
\begin{enumerate}
    \item The implementation of a Kolmogorov-Arnold Network model for recursive learning of deep-Koopman (RLDK) that is comparably more efficient and accurate than its MLP counterpart. 
    \item A comparison of KANs with that of an MLP network on first the pendulum dynamics to prove it's the capability and to demonstrate a controllable system, and second on the dynamics of the two-body problem to emphasize the applicability on a real-world system.
    \item Quick and efficient development of Koopman operator through KANs for the fully under actuated pendulum-cart dynamics, with successful real-time implementation of LQR control on the nonlinear system.
\end{enumerate}
Section \ref{sec:II} provides preliminary information regarding the Koopman operator and its derivation, EDMD/EDMDc and how it is used in the formulation of the Koopman operator, and finally, the KANs architecture which is the main development of this paper. The pendulum and two-body problem used to compare the two networks is presented in Section \ref{sec:III} along with the pendulum-cart system, followed by results and discussions. Section \ref{sec:IV} concludes this paper. 

\section{Preliminaries and Theory} \label{sec:II}
\subsection{Koopman Theory Preliminaries}

Koopman operator theory, originally proposed in 1931 by B.O Koopman \cite{Koopman1931-ev} defines the necessary method to map any nonlinear dynamical system to an infinite-dimensional linear system. For the purpose of completeness in this letter, we show the implementation of Koopman Theory through EDMDc. Although the 2BP system is an uncontrolled system, the control input in both the data generation and learning framework can be set to zero. This results in straight EDMD as the data driven method for the discovery of the Koopman operator. Suppose we have a controlled discrete-time nonlinear dynamical system defined as

\begin{equation} \label{nonlineardyn}
    x_{k+1} = \boldsymbol{f}(x_k, u_k),
\end{equation}

where \(x_k \in \mathcal{M} \subset \mathbb{R}^n \) and \(u_k \in \mathbb{R}^p \) is the system state and control input respectively, \(k\) defines the index for each time step, and \(\boldsymbol{f}\) is the function that evolves the states through state space. We can then define observable functions, which here are real-valued square-integrable functions of the system state: \(g : \mathbb{R}^n \rightarrow \mathbb{R}\). For any observable function $g$, the discrete time Koopman operator, $\mathcal{K}_{\Delta t}$, can then be defined as,
\begin{equation}
    \mathcal{K}_{\Delta t}g = g\circ \boldsymbol{f},
\end{equation}

where \(\circ\) is the composition operator. We can now apply this operator to the discrete-time system to arrive at:

\begin{equation}\label{eqn:koop}
    \mathcal{K}_{\Delta t}g(x_k) = g(\boldsymbol{f}(x_k)) = g(x_{k+1}).
\end{equation}

Equation \ref{eqn:koop} shows that the Koopman operator propagates an observable function of any state, \(g(x_k)\), to the next time step. For the sake of clarity, all further reference to the Koopman operator will drop the $\Delta t$, whilst still referring to the discrete-time operator.

The theory presented above implies that the lifted linear system is of infinite dimension, an impractical assumption. Hence, a great deal of research has been conducted to investigate methods to select an adequate finite-dimensional Koopman observable space \cite{Mezic2005-oa,Lusch2018-oh,Xiao2023-ml}. Since this task of selecting observable functions is inherently very complex and involved, we utilize EDMD and its extensions\cite{Williams2015-zc} to formulate a finite-dimensional approximation of the Koopman operator. EDMD involves lifting time-series trajectory data into higher-dimensional space through a set of basis observable functions and applying them to each $x_k$. Then, a linear time invariant (LTI) matrix, $\mathcal{K}$, can be fit to apply to the higher-dimensional system. Observable functions are commonly built from a set of basis functions such as monomials, higher-order polynomials, radial basis functions, trigonometric functions, etc. \cite{Mezic2005-oa,Folkestad2020-yk}. We refer readers to Brunton et al. \cite{Brunton2021-do} and \cite{Brunton2022-ya} for an in-depth study of Koopman operator theory.

\subsection{Koopman Algorithm}

In this work, a comparison of performance between a KANs and MLP deep neural network (DNN) is conducted. Both utilize EDMD and EDMDc to approximate the Koopman operator in finite dimensions. First, rather than hand-selecting a set of observable functions, the DNN defines the set of observable functions. The MLP achieves this through its weights, biases and activation functions on the nodes, whilst KANs achieves this through the spline activation between the nodes \cite{Liu2024-dx}. Then, the Koopman operator is calculated using least-squares regression on the custom loss function defined in Section \ref{NN Arc}.

Given the nonlinear system defined in Equation \ref{nonlineardyn}, we can apply the observable mapping, represented by the operator $\boldsymbol{\Phi}$, to the states to map them to the higher lifted space. We then attempt to use the DNN to find a linear representation of the system:

\begin{equation} \label{LTIsys}
    \mathbf{\Phi}(x_{k+1}) \approx \mathbf{K} \boldsymbol{\Phi}(x_k) + \mathbf{B}\mathbf{u}_k,
\end{equation}

where the matrix $\mathbf{K}$ is the approximated Koopman operator analogous to the LTI system $\mathbf{A}$ matrix, while $\mathbf{B}$ is the input matrix. We choose \(N\) such that \(N > n\) and define

\begin{equation}\label{eqn:observ}
    \mathbf{\Phi}(\textbf{x}_k) := 
    \begin{bmatrix} 
    \textbf{x}_k \\
    \phi_1(\textbf{x}_k) \\ \phi_2(\textbf{x}_k) \\ \vdots \\ \phi_N(\textbf{x}_k) 
    \end{bmatrix},
\end{equation}

where \(\mathbf{\phi}_i : \mathbb{R}^n \rightarrow \mathbb{R}, i = 1,...,N\) are the observable functions, learned by the DNN. 

Note that we concatenate the original states \(\mathbf{x}_k\) with the observable. This helps by allowing easy extraction of the original states from the observables, ideal for control development.
. 
The choice of the size of$N$ is an important parameter, but there is currently no one method for analytically determining its size; therefore, in most cases, \(N\) is chosen empirically. Whilst the number of observables is important, finding a 'good' choice of basis functions to create the observable functions leads to a more accurate approximation of the operator. 
Some work has been done to study the optimal size $N$ to find controllable systems \cite{Zinage2023-hg,Lusch2018-oh}. In the case of KANs, in this work, the size and observables are determined by the nonlinear splines that connect each node between the hidden layers. 

To calculate the approximate Koopman operator $\mathbf{K}$ and the input matrix $\mathbf{B}$, the state time history data for \(M\) steps is arranged into what are known as snapshot matrices. The first snapshot matrix, \(X\), is the state history from time \(k=1\) to \(k=M-1\), whilst the second matrix, \(X'\) is the same state history, right shifted by one-time step and the third snapshot matrix $U$ is of the control history such that: 

\begin{equation} \label{eqn:snapX}
    X = \begin{bmatrix} x_{1},  x_{2},  x_{3},  \dots, x_{M-1} \end{bmatrix} 
\end{equation}
\begin{equation} \label{eqn:snapY}
    X' = \begin{bmatrix} x_{2},  x_{3}, x_{4},  \dots, x_{M} \end{bmatrix} 
\end{equation}
\begin{equation} \label{eqn:snapU}
    U = \begin{bmatrix} u_{1},  u_{2}, u_{3},  \dots, u_{M-1} \end{bmatrix} 
\end{equation}

Mapping the measured state data with the observable functions leads to:

\begin{equation}
    \boldsymbol{\Phi}(X) = \begin{bmatrix} \boldsymbol{\Phi} (x_1), \boldsymbol{\Phi}(x_2),  \dots, \boldsymbol{\Phi}(x_{M-1}) \end{bmatrix} 
\end{equation}
\begin{equation}
    \boldsymbol{\Phi}(X') = \begin{bmatrix} \boldsymbol{\Phi}(x_2), \boldsymbol{\Phi}(x_3),  \dots, \boldsymbol{\Phi}(x_M)  \end{bmatrix} 
\end{equation}    

Given this dataset, the matrices $\mathbf{K}$ and $\mathbf{B}$ can be found by solving the least-sqaures problem:

\begin{equation}
\min \sum\left\|\boldsymbol{\Phi}(x_{k+1})-\left(\mathbf{K} \mathbf{\Phi}\left(x_k\right)+\mathbf{B} u_k\right)\right\|^2.
\end{equation}

Applying the snapshot matrices of real system data yields:

\begin{equation}
    \mathbf{\Phi(X')} \approx 
    \mathbf{K} \mathbf{\Phi(X)}+ \mathbf{B}\mathbf{U} = \begin{bmatrix}
        \mathbf{K} & \mathbf{B}
    \end{bmatrix}
    \begin{bmatrix}
        \mathbf{\Phi(X)} \\ \mathbf{U}
    \end{bmatrix};
\end{equation}

therefore, 

\begin{equation} \label{eqn:EDMDc}
    \begin{bmatrix}
        \mathbf{K} & \mathbf{B}
    \end{bmatrix} = \boldsymbol{\Phi(X')}\begin{bmatrix}
        \mathbf{\Phi(X)} \\ \mathbf{U}
    \end{bmatrix}^\dagger,
\end{equation}

where $\dagger$ denotes the Moore-Penrose inverse of the matrix \cite{Penrose1955-cj}. 

Because the Koopman operator calculated with least-squares is an approximation, and as the observable functions do not span a Koopman invariant subspace, the predicted state is an approximation of the real state, thus we denote it with the $\hat{}$ symbol:

\begin{equation}
    \hat{\mathbf{\Phi}}(x_{k+1}) = \mathbf{K} \hat{\mathbf{\Phi}}(x_k) + \mathbf{B} \mathbf{u}_k.
\end{equation}

Now, we can extract the original states from the observables using a projection matrix $\mathbf{P}$ \cite{Junker2022-qh} yielding

\begin{equation} \label{eqn:extract}
    x_{k+1} = \mathbf{P}\hat{\mathbf{\Phi}}(x_{k+1})\; \text{with}\; \mathbf{P} = \begin{bmatrix}
        \mathbf{I}_n, \mathbf{0}_{n \text{x} N}
    \end{bmatrix},
\end{equation}

where \(\mathbf{I}_n\) is the \(n\) x \(n\) identity matrix and \(\mathbf{0}_{n \text{x} N}\) is the \(n\; \text{x}\; N\) zero matrix. As shown in \cite{Junker2022-qh} and \cite{Korda2018-qs}, the observable functions not spanning a Koopman invariant subspace leads to an accumulation of error over time, which can lead to misleading predictions. However, if the predicted state is corrected at each time step, this error can be mitigated. This correction is applied by extracting the estimated state variable $\hat{x}_{k+1}$ at each time step with Equation \ref{eqn:extract} and then reapplying the observable mapping to the extracted state variable with Equation \ref{eqn:observ} to result in the next state. 

\begin{figure}[ht]
    \centering
    \includegraphics[width = \columnwidth]{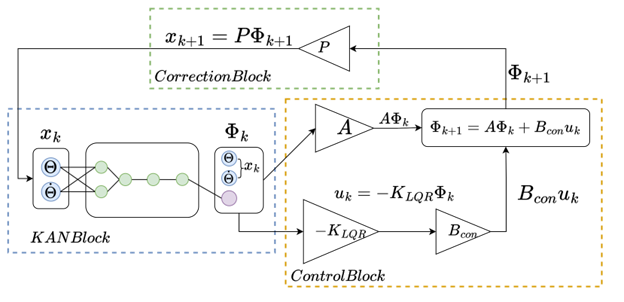}
    \caption{Complete schematic of dynamic propagation of the states, including the control input. Note that the KANs block is very simple and non-complex}
    \label{fig:1}
\end{figure}

Figure \ref{fig:1} shows the proposed architecture for the trained network with prediction and control in continuous time. Here, the given initial states pass through the KAN DNN block to form the lifted states. The lifted states are then concatenated with original states $\mathbf{x}_k$ to form the new set of observables $\mathbf{\Phi}$. The states are propagated through the control block where $\mathbf{A}$ and $\mathbf{B}_{con}$ are the continuous form of Koopman matrix $\mathbf{K}$ and $\mathbf{B}$. The predicted states then pass through the correction block, resulting in the original states' extraction. The entire loop runs until the states are regulated. 

\subsection{Neural Network Architecture} \label{NN Arc}

The main contribution and improvement of this paper is the implementation of the KANs architecture as the DNN responsible for learning the Koopman operator. The benefit of using KANs over MLP is that the KANs network is more accurate and parameter efficient than MLP DNN's. KANs is able to achieve this because it uses 1D functions parametrized as splines as the activation functions between nodes, with the nodes simply containing a summing operation. 

MLP's take advantage of the universal approximation theorem (UAT) \cite{Cybenko1989-sd}, which states that for a given continuous function and a given error $\epsilon > 0$, a two-layer network, with neurons $ n > O(\epsilon^{\frac{-2}{m}})$, where $m$ is the order up to which the function is continuously differentiable, is able to approximate that function within that error. The issue however, is that UAT can often, for complex nonlinear functions, require a very high dimensional representation of the function. 

The Kolmogorov Arnold Theorem (KAT) reveals that for any multivariate continuous function $f$ bounded by a domain, it can be represented as a composition of finite linear sums of a number of univariate continuous functions. The mathematical representation of this theorem is as such:

\begin{equation} \label{eqn:KAT}
    f(\mathbf{x}) = f(x_1,x_2,...,x_n) = \sum_{q=1}^{2n+1} \mathbf{\Psi}_q \left( \sum_{p=1}^{n} \psi_{q,p}(x_p) \right)
\end{equation}

where $f : [0,1]^n \rightarrow \mathbb{R}$ is a smooth function, $\psi_{q,p} : [0,1] \rightarrow \mathbb{R}$ and $\mathbf{\Psi}_{q} : \mathbb{R} \rightarrow \mathbb{R}$. The formulation given in Equation \ref{eqn:KAT} shows what a $2n+1$ layered network would look like, but the work Liu et.al \cite{Liu2024-dx} demonstrates what it would look like to have a KANs with arbitrary width and depth. 

For clarity, we briefly explain the architecture of KANs and how it treats nonlinearities differently to MLPs. We can define the shape of a KAN by the integer array 

\begin{equation}
    [n_0,n_1,...m_L]
\end{equation}

where $n_i$ is the number of neurons in the $ith$ layer of the network of total layers $L$. We then define the mathematical representation of a layer within the KAN as the inner sum of Equation \ref{eqn:KAT}, where each $\psi_{q,p}$ is the activation function between the neuron in the previous and current layer. We now see that the composition of all neuron activations across the layers can be represented with $\mathbf{\Psi}_p$, hence the general KANs network for $L$ layers can be described by:

\begin{equation} \label{KANeqn}
    KAN(\mathbf{x}) = (\mathbf{\Psi}_{L-1}\circ\mathbf{\Psi}_{L-2}\circ\dots\circ\mathbf{\Psi}_{1}\circ\mathbf{\Psi}_{0})\mathbf{x}
\end{equation}

Expanding these compositions through the layers, to show the summation of the activations within each layer gives:

\begin{multline}
    f(\mathbf{x}) = \sum_{i_{L-1}=1}^{n_{L-1}} \psi_{L-1,i_L,i_L-1} \Bigg( \Bigg. \sum_{i_{L-2}=1}^{n_{L-2}} \dots \\ \left( \sum_{i_1=1}^{n_1} \psi_{1,i_2,i_1} \left( \sum_{i_0=1}^{n_0}\psi_{0,i_1,i_0}(x_{i0}) \right) \right) \dots \Bigg. \Bigg)
\end{multline}

where $\psi_{l,i,j}$ is the activation function in layer $l$ between the $jth$ neuron in the $lth$ and the $ith$ neuron in the $(l+1)$ layer.

For a comparison, an MLP, which deals with the nonlinearities in its activation function $\sigma$ and linear transformations in $\mathbf{W}$, can be represented as:

\begin{equation} \label{MLPeqn}
    MLP(\mathbf{x}) = (\mathbf{W}_{L-1}\circ\sigma\circ\mathbf{W}_{L-2}\circ\sigma\circ\dots\circ\mathbf{W}_{1}\circ\sigma\circ\mathbf{W}_{0})\mathbf{x}
\end{equation}

It is clear from Equation \ref{KANeqn} that the KANs deals with both nonlinearities and transformations together in $\mathbf{\Psi}$ whilst the MLP treats them separately in $\mathbf{W}$ and $\sigma$ as shown in Equation \ref{MLPeqn}.

We refer the reader to Liu et. al \cite{Liu2024-dx} for a more extensive description on KANs and more information on hyperparameters, training and examples.

\subsection{Loss Function}

In order to achieve accurate representation of the Koopman operator, the DNN is trained using a custom loss function that is similar for both KANs and the MLP network to allow for an accurate comparison. The DNN first learns the observable functions which are in turn used to learn the Koopman operator through EDMD or EDMDc. The loss function is a weighted summation of 2 separate loss functions that act to improve the approximation in different ways. The first, being the reconstruction loss function: 

\begin{equation} \label{eqn:recon_loss}
\mathcal{L}_{\text {Recon}}=\frac{1}{N_d} \sum_{k=1}^{N_d}\left\|{\hat{\boldsymbol{x}}_{k+1}}-{\boldsymbol{x}_{k+1}} \right\|_2^2
\end{equation}

where $\hat{\boldsymbol{x}}_{k+1}$ is the one time-step predicted state defined in Equation \ref{eqn:extract}, and $\mathbf{x_{k+1}}$ is the corresponding state calculated using the true data. The purpose of this loss function is to ensure the reconstruction of the states from the lifted space is accurate.

If the reconstruction loss was the sole loss function implemented, then as time evolved in the prediction, error would propagate resulting in a drift of the predicted linear dynamics from the true nonlinear dynamics. Hence, we introduce a second loss function, the prediction loss, which aims to improve the prediction over time, by comparing the extracted state not only one time-step in the future, but multiple. The length of which is determined by the user in the data generation phase and given by $\alpha$. 

\begin{equation} \label{eqn:pred_loss}
    \mathcal{L}_{\text {Pred}}=\frac{1}{N_{pred}} \sum_{k=1}^{N_{pred}}\left\|{\hat{\boldsymbol{x}}_{k+\alpha}}-{\boldsymbol{x}_{k+\alpha}} \right\|_2^2
\end{equation}

The total loss function can then be calculated as a weighted sum of the two loss functions, where the weights $\gamma$ and $\beta$ are chosen as hyperparameters of the training:

\begin{equation}
    \mathcal{L}_{\text {total}}=\gamma\mathcal{L}_{\text {Pred}}+\beta\mathcal{L}_{\text {Recon}}
\end{equation}

The total loss is then calculated by recursively applying the mean squared error (MSE). It is noted, that the MLP loss function utilizes L1 and L2 regularization to enforce sparsity in the model and reduce over and under fitting, however the KANs model does not include these added regularizations. L1 regularization is defined for KANs \cite{Liu2024-dx} but requires an added entropy loss to be added alongside. This is left for future work.

The full PyTorch code, including the data generation, training and simulations are on Github\footnote[1]{\url{https://github.com/tiwari-research-group/Koopman-with-KANs}}. 

\section{Simulation and Results}
\label{sec:III}

In this section, we present the simulation, results, and discussion of the RLDK method with KANs. The proposed methodology is also compared against a traditional MLP DNN architecture for two of the systems, with a third, more complex system used to highlight the effectiveness of this methodology. The two dynamic systems used for comparison are the pendulum and the Two-Body problem and are simulated as shown in the loop Figure \ref{fig:1}. Because this is an extension, and the main goal of this work is to highlight the use of KANs in the context of Koopman Theory, the full explanation of these systems can be found in our previous works \cite{Tiwari2023-ya, Nehma2024-fi}. The final system is the pendulum-cart system which is often used to demonstrate the ability to control a highly nonlinear and fully under actuated system. The purpose is to challenge KANs and our methodology to develop an accurate linear global approximation of the nonlinear system so that we can build an sufficient LQR controller that works well on the nonlinear system.

\subsection{Simulation Setup}

A key note regarding the simulation and training of all of the models is that the entire framework for the KANs was carried out on an 11th Gen Intel Core i7 CPU. This is large contrast to all traditional MLP DNN frameworks which often require expensive, and demanding GPU's to perform the training. As such all training and simulations for the MLP DNN was conducted with an NVIDIA GeForce RTX 3090 GPU and this fact should be kept in mind when considering training times and performances.

\subsubsection{Pendulum}
The pendulum is a common example that is widely used due to its relatively simple yet nonlinear dynamics. The state dynamics are given as follows:

\begin{equation}
    \begin{bmatrix} x_1 \\ x_2
    \end{bmatrix} = 
    \begin{bmatrix} \theta \\ \Dot{\theta} 
    \end{bmatrix}
    \label{pend 1}
\end{equation}

\begin{equation}
    \Dot{x} = \begin{bmatrix} \Dot{x_1} \\ \Dot{x_2}
    \end{bmatrix} = 
    \begin{bmatrix} x_2 \\ -\frac{g}{l}sin(x_1) 
    \end{bmatrix}
    \label{pend 2}
\end{equation}

For the training data used in the KANs model, the pendulum dynamics were integrated for 2 seconds using a Runge-Kutta integrator ($\Delta t = 0.01sec$), using 10 (8000 for the MLP) random initial conditions within \(-2 \leq \theta_0 \leq 2\) rad and \(-2 \leq \dot{\theta}_0 \leq 2\) rad/s. For each initial condition, the state history, the state history shifted by one time step, the state history shifted by $\alpha$ time steps and the control history (randomly generated in the range $[-0.1,0.1]$ were captured. All four datasets were arranged into snapshot matrices as given in Equations \ref{eqn:snapX}, \ref{eqn:snapY} and \ref{eqn:snapU}.

\subsubsection{Two-Body Problem}
To compare the two networks on a more complex and applicable system, the 2BP was chosen. The state dynamics are given as follows:

\begin{equation}
    \mathbf{x} = \begin{bmatrix} x_1 \\ x_2 \\ x_3 \\ x_4
    \end{bmatrix} = 
    \begin{bmatrix} x \\ y \\ \dot{x} \\ \dot{y}
    \end{bmatrix}
    \label{orbit eom}
\end{equation}

\begin{equation}
\begin{aligned}
    \dot{\mathbf{x}} &= \begin{bmatrix} \dot{x_1} \\ \dot{x_2} \\ \dot{x_3} \\ \dot{x_4}
    \end{bmatrix} = 
    \begin{bmatrix} x_3 \\ x_4 \\ \frac{-\mu\cdot \vec{x_1}}{\sqrt{x_1^2 + x_2^2}^3}\\  \frac{-\mu \cdot \vec{x_2}}{\sqrt{x_1^2 + x_2^2}^3}
    \end{bmatrix}
    \label{orbit ss}
\end{aligned}
\end{equation}

To ease in the development of training data, the initial conditions of the 2BP was set to be the periapsis of the orbit, hence $x_0 = r, \dot{x}_0 = 0, y = 0, \dot{y}_0 = \sqrt{\frac{\mu}{r}}$. Where $r$ is the radius of the orbit and is randomly generated in the range $[6578, 11378]$ km. For the KANs model, 30 initial conditions were generated for the training data (100 for the MLP network) and an $\alpha=15$ ($\alpha=25$ for MLP). Each initial condition was propagated for one orbit with each data set containing 800 data points. Like the pendulum problem the data was organised into the snapshot matrices minus the control snapshot matrix as these dynamics did not include a control input.  

\subsubsection{Pendulum Cart}
The pendulum-cart is fully under actuated system, with only one control input and two degrees of freedom and is often used to validate the applicability of nonlinear controllers. In this work we use its complex nonlinearities to challenge our architecture to control this meaningful example. More detailed explanations of the this system can be found in \cite{pend-cart1,pend-cart2}. The state dynamics are given by: 

\begin{equation}
    \mathbf{x} = \begin{bmatrix} x_1 \\ x_2 \\ x_3 \\ x_4
    \end{bmatrix} = 
    \begin{bmatrix} x \\ \theta \\ \dot{x} \\ \dot{\theta}
    \end{bmatrix}
    \label{pend-cart eom}
\end{equation}

\begin{equation}
\begin{aligned}
    \dot{\mathbf{x}} &= \begin{bmatrix} \dot{x_1} \\ \dot{x_2} \\ \dot{x_3} \\ \dot{x_4}
    \end{bmatrix} = 
    \begin{bmatrix} x_3 \\ x_4 \\ \frac{u + m_p\sin{x_2}(l\dot{x_2}^2-g\cos{x_2})}{m_c + m_p\sin^{2}{x_2}}  \\ \frac{u\cos{x_2} + m_pl\dot{x_2}^2\cos{x_2}\sin{x_2}-(m_c+m_p)g\sin{x_2}}{l(m_c+m_p\sin^{2}{x_2})}
    \end{bmatrix}
    \label{pend-cart ss}
\end{aligned}
\end{equation}

Where $u \in \mathbb{R}$ is the control input that corresponds to the acceleration of the cart on the track, $l$ is the length of the pendulum, $g$ is the value of gravity and, $m_c$ and $m_p$ are the masses of the cart and pendulum respectively. For the generation of training data, the initial conditions for each of the states were randomly selected in the range $[-1,1]$. 30 trajectories were generated for use as the training data with each trajectory run for 15 seconds with a timestep, $dt$ of 0.1 seconds and $\alpha = 10$. The system was simulated with random control excitation in the range $[-0.08,0.08]$ at each time step, thus the state and control history data could be arranged in snapshot matrices as outlined previously. 
\subsection{Results and Discussions}
\label{sec:IV}

To best summarise the difference between the KANs model and the MLP model, we organised key properties of the networks and their properties into Tables \ref{table:pend_net} and \ref{table:2bp_net} which can be found in \ref{app1}.

\begin{figure}[hbt!]
    \centering
    \includegraphics[width=\columnwidth]{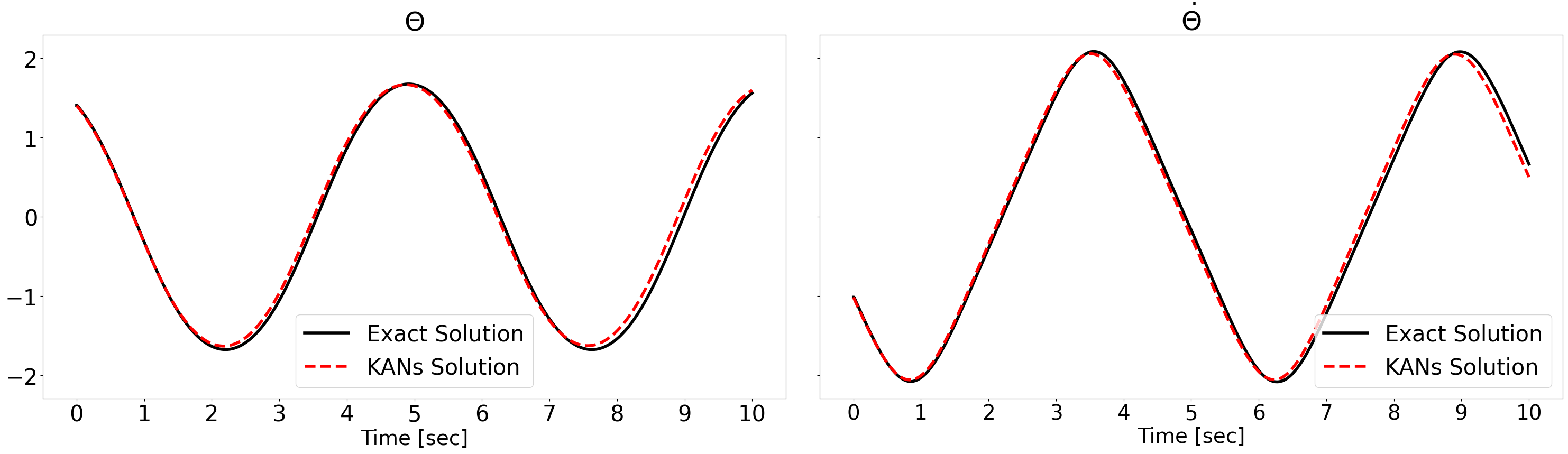}
    \caption{KANs learned, self-propagating dynamics prediction given only the same initial condition as the ground truth nonlinear dynamics}
    \label{fig:KAN_pend}
\end{figure}

\begin{figure}[hbt!]
    \centering
    \includegraphics[width=\columnwidth]{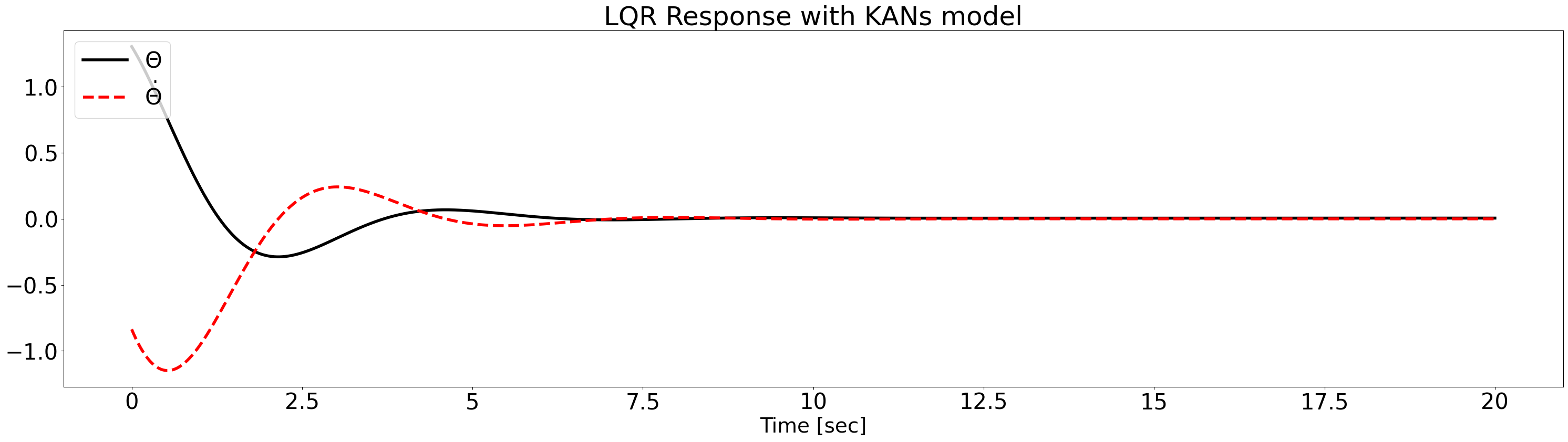}
    \caption{System response to LQR Controller developed using the linear system created with the KANs learned Koopman operator.}
    \label{fig:KAN_LQR}
\end{figure}

From Table \ref{table:pend_net} it is clear that KANs is significantly smaller than that of the MLP network, by about half. The discovered Koopman operator is smaller, of size 3x3, but the main difference between the models is the fact that KANs is able to accurately learn the operator with one hidden layer of size 1. This leads to a decrease in the training time required to achieve the results displayed in Figure \ref{fig:KAN_pend}. This smaller network and Koopman operator size result in a lower burden for the onboard computer utilizing this linearization technique, which for many applications is greatly beneficial. Another of the crucial benefits that KANs provide is that it is able to complete its learning on a significantly smaller data set, in that the MLP required 8000 different trajectories, whilst KANs only needed 15 trajectories to learn the dynamics. This is a tremendous benefit when applying to real-world systems, as collected data may be scarce or not abundant enough to support traditional MLP network training. This also opens the possibility to online training to further improve the model or increase the state space domain in which it accurately represents the nonlinear system. Despite training faster, being smaller parameter-wise, and utilizing less data, KANs were also able to achieve 1.25 times greater accuracy in the prediction of the true dynamics as compared to the prediction of the MLP network. As seen in Figure \ref{fig:KAN_LQR}, the model learned by KANs is still able to be used with an LQR controller to regulate the states within a reasonable period, showing that the utilization of EDMDc within the loss function was correct. 

\begin{figure}[hbt!]
    \centering
    \includegraphics[width=0.9\columnwidth]{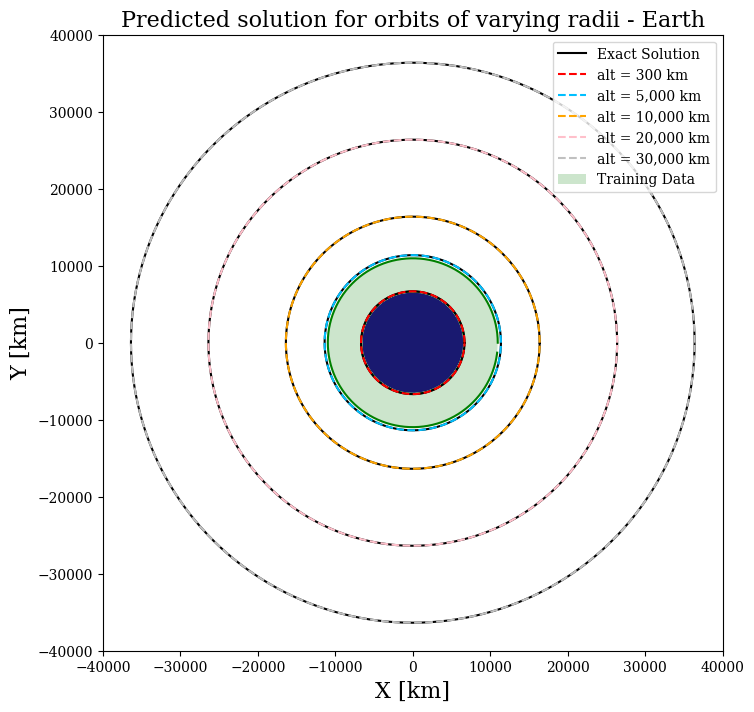}
    \caption{KANs learned, linear propagation of multiple orbits of varying altitudes.}
    \label{fig:KAN_2BP}
\end{figure}

The benefits of KANs over MLP networks is more evident in the case of the 2BP. From Table \ref{table:2bp_net}, we can see the significant decrease in parameter size of the network, which leads to a dramatic reduction in training time, from $\sim$47 minutes to under 1.5 minutes. Not only this but the number of training trajectories is also shaved down greatly. The size of the Koopman operator needed is also smaller for KANs, highlighting the fact that the learned observable functions are better approximations of the Koopman eigenfunctions. With these significant increases in efficiency and reduction in size, KANs are still able to generate results comparable to those of the MLP network in terms of maximum absolute error. Figure \ref{fig:KAN_2BP} demonstrates the capability of the KANs learned Koopman model to accurately predict a range of orbits, extending beyond the original training range.  

\begin{figure}[hbt!]
    \centering
    \includegraphics[width=\columnwidth]{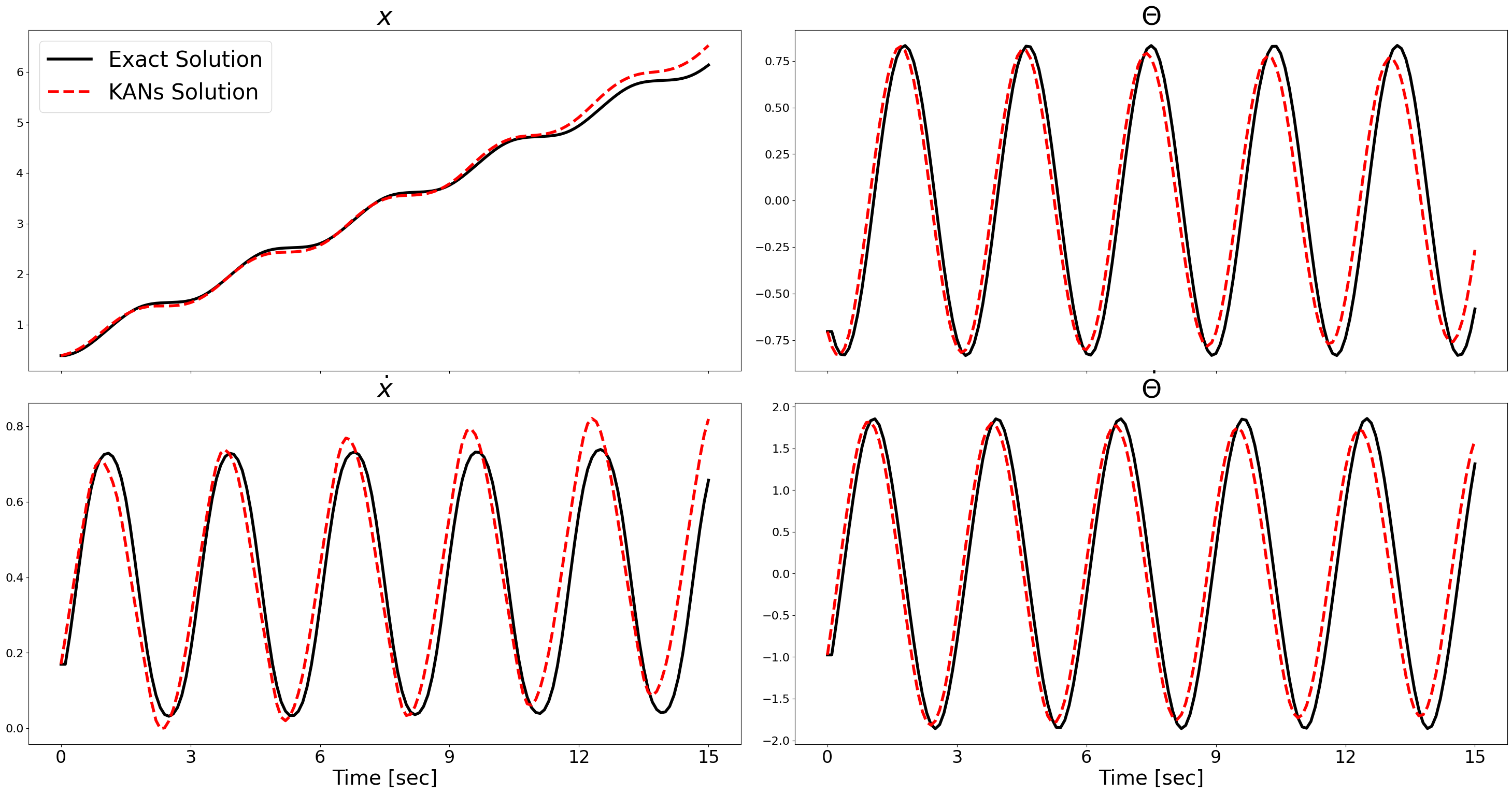}
    \caption{KANs learned, linear propagation of pendulum-cart system with random excitation as the control input.}
    \label{fig:pendcart_dyn}
\end{figure}

\begin{figure}[hbt!]
    \centering
    \includegraphics[width=0.7\columnwidth]{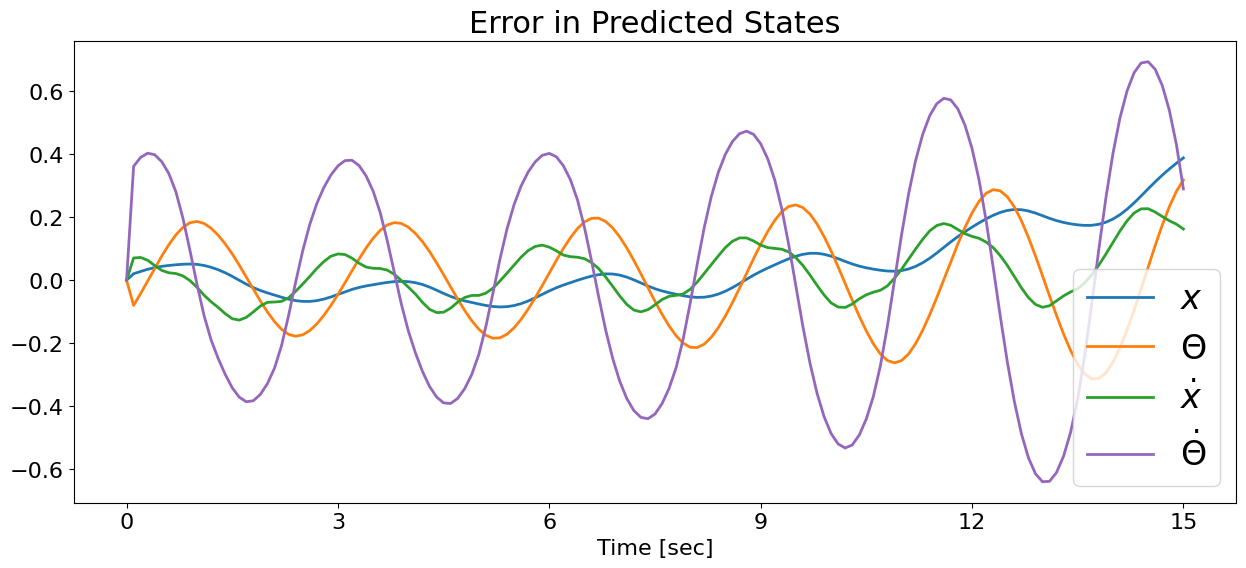}
    \caption{Error between the true nonlinear dynamics and the linear approximation from the Koopman operator.}
    \label{fig:pendcart_error}
\end{figure}

The pendulum-cart model also exhibited favorable results. Figure \ref{fig:pendcart_dyn} depicts the evolution of the system states between the linear and nonlinear models where it is clear that the linear time-invariant Koopman model closely follows the nonlinear model for the duration of the simulation. As the system evolves, the error gradually grows, most evident in Figure \ref{fig:pendcart_error} where the error in the $x$ and $\dot{x}$ states increasingly growing faster than the other states. The prediction error growing through time is an expected result due to the fact that the Koopman operator learned by KANs is a numerical approximation of the true Koopman operator. A true Koopman operator, however, would be expected to have zero error. The training time for this model was $\sim 6.5 minutes$ which is relatively fast and emphasises the potential for online learning through KANs, whilst the total number of parameters for the model is 386.

\begin{figure}[hbt!]
    \centering
    \includegraphics[width=\columnwidth]{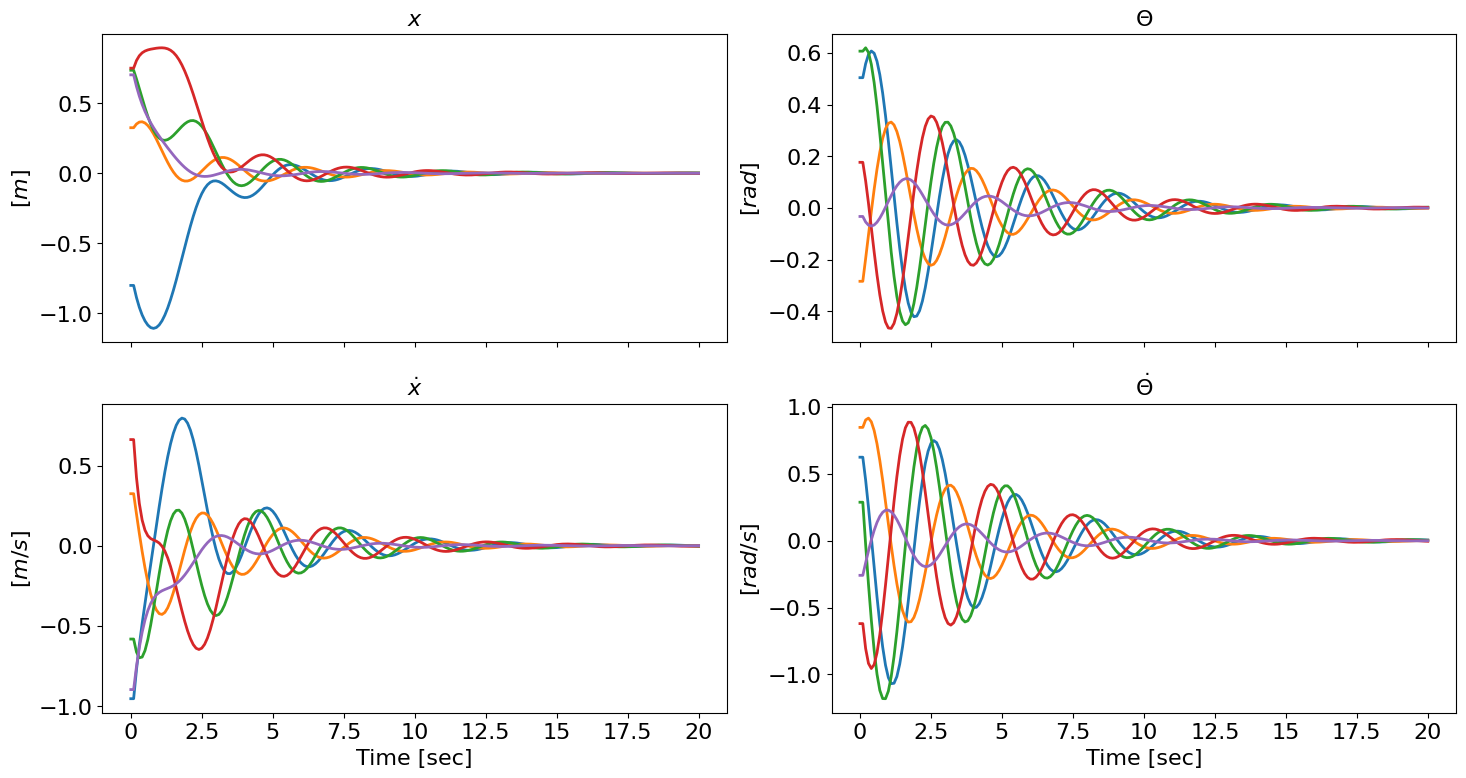}
    \caption{State history for 5 random trajectories with the LQR controller applied. The control input, designed on the linear system, is applied to the nonlinear dynamics.}
    \label{fig:pendcart_cont}
\end{figure}

\begin{figure}[hbt!]
    \centering
    \includegraphics[width=0.6\columnwidth]{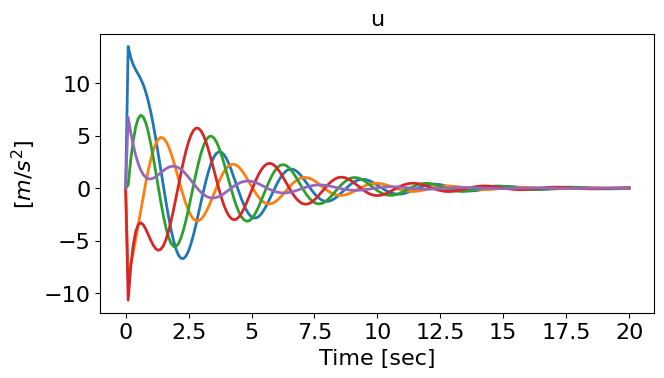}
    \caption{Control history for the same 5 trajectories simulated above.}
    \label{fig:lqr_cont}
\end{figure}

As with the pendulum system described earlier, an LQR controller was designed to showcase the ability of this linear model to be used in the development of a controller for the nonlinear system. The LQR gains were tuned on the Koopman linear model and then fed back to the nonlinear system resulting in the regulation of all the states as seen in Figure \ref{fig:pendcart_cont}. We show that a controller developed with the Koopman model is able to accurately control the nonlinear system with good performance whilst also exhibiting realistic and achievable control demands as seen in Figure \ref{fig:lqr_cont}. 

\section{Conclusions} \label{sec:VI}
In this work, the KANs framework was successfully used to develop a more efficient, faster and accurate linear Koopman approximation of the pendulum,Two-Body and pendulum-cart nonlinear dynamics. Comparing KANs to the previous MLP model for both the pendulum and 2BP systems shows that the smaller and less computationally expensive KANs can perform similarly, if not better than the MLP model despite requiring less data, shorter training times, and removing the need for a GPU. We demonstrate the capabilities of the this methodology with KANs to quickly and effectively learn and be applied to a fully under actuated system in the pendulum-cart system, and demonstrate the ability to design an LQR controller capable of achieving the control of the nonlinear system. We show that KANs is a valid and promising framework for the future development of deep Koopman operators and hence their applicability to real-time, real-world applications and use cases. Although KANs is highly interpretable in learning nonlinear functions, this does not translate well to a model learning Koopman Theory. We leave the prospect of training improvements such as L1 and L2 regularization to future work. 

\clearpage
\appendix
\section{Comparison Tables between the MLP networks and KANs}
\label{app1}

\begin{table}[!htbp]
\begin{center}
\begin{tabular}{ |p{6cm}|p{2.5cm}|p{2.5cm}| }
 \hline
 \multicolumn{3}{|c|}{\textbf{Pendulum DNN Comparison}} \\
 \hline
 \textbf{Parameter} & \textbf{MLP} & \textbf{KANs} \\
 \hline
 Lifted Space Size & 4 & 3 \\
 Hidden Layers & 8 & 1\\
 Neurons per hidden layer & 6 & 1\\
 Total Parameters & 326 & 59 \\
 Batch Size & 4096 & N/A\\
 Learning Rate & 0.0001 & 1\\
 Optimizer &   Adam & LBFGS\\
 Activation Function & SELU & B Spline\\
 Weight Decay & 0.00001 & N/A\\
 Epochs & 10000 & 3\\
 $\alpha$ & 25 & 25 \\
 \(\gamma\) & 0 & 1\\
 \(\beta\) & 1 & 1\\
 \(\lambda_{L_1}\) & 0 & N/A\\
 \(\lambda_{L_2}\) & 0 & N/A\\
 \hline
 K matrix dimension & 4 x 4 & 3 x 3\\
 \hline
 Training Time & 30 sec & 15 sec \\
 Training Data Required & 8000 IC & 15 IC \\
 Training Platform & RTX 3090 & Intel Core i7 \\
 Max Absolute Error & 0.2 rad & 0.15 rad\\
 \hline
\end{tabular}
\end{center}
\caption{Comparison of hyperparameters and performance between KANs and MLP for Pendulum Problem}
\label{table:pend_net}
\end{table}

\begin{table}[!htbp]
\begin{center}
\begin{tabular}{ |p{6cm}|p{2.5cm}|p{2.5cm}| }
 \hline
 \multicolumn{3}{|c|}{\textbf{Two-Body DNN Comparison}} \\
 \hline
 \textbf{Parameters} & \textbf{MLP} & \textbf{KANs} \\
 \hline
 Lifted Space Size & 10 & 5 \\
 Hidden Layers & 3 & 3\\
 Neurons per hidden layer & 25 & 1\\
 Total Parameters & 1581 & 102 \\
 Batch Size & 1 & N/A\\
 Learning Rate & 0.0001 & 0.0001\\
 Optimizer &   Adam & LBFGS\\
 Activation Function & SELU & B Spline\\
 Weight Decay & 0.00001 & N/A\\
 Epochs & 80000 & 10\\
 $\alpha$ & 15 & 15 \\
 \(\gamma\) & 0.8 & 1\\
 \(\beta\) & 1 & 1\\
 \(\lambda_{L_1}\) & 0.04 & N/A\\
 \(\lambda_{L_2}\) & 0.01 & N/A\\
 \hline
 K matrix dimension & 10 x 10 & 5 x 5\\
 \hline
 Training Time & 47 min & 1.5 min \\
 Training Data Required & 200 IC & 30 IC \\
 Training Platform &RTX 3090 & Intel Core i7 \\
 Max Absolute Error & 3 km & 2.4km \\
\hline
\end{tabular}
\end{center}
\caption{Comparison of hyperparameters and performance between KANs and MLP for 2BP}
\label{table:2bp_net}
\end{table}

\clearpage
\bibliographystyle{elsarticle-harv}
\bibliography{bibliography.bib}

\end{document}